\pdfoutput=1
\documentclass[conference]{IEEEtran}
\IEEEoverridecommandlockouts
\usepackage{cite}
\usepackage{amsmath,amssymb,amsfonts}
\usepackage{algorithmic}
\usepackage{graphicx}
\usepackage{textcomp}
\usepackage{xcolor}
\usepackage{subfigure}
\def\BibTeX{{\rm B\kern-.05em{\sc i\kern-.025em b}\kern-.08em
		T\kern-.1667em\lower.7ex\hbox{E}\kern-.125emX}}
\begin{document}
	
	\title{Multiple Linear Regression Haze-removal Model Based on Dark Channel Prior}
	
	\author{\IEEEauthorblockN{1\textsuperscript{st} Binghan Li}
		\IEEEauthorblockA{\textit{Electrical and Computer Engineering} \\
			\textit{Texas A\&M University}\\
			College Station, TX \\
			lbh1994usa@tamu.edu}
		\and
		\IEEEauthorblockN{2\textsuperscript{nd} Wenrui Zhang}
		\IEEEauthorblockA{\textit{Electrical and Computer Engineering} \\
			\textit{Texas A\&M University}\\
			College Station, TX \\
			zhangwenrui@tamu.edu}
		\and
		\IEEEauthorblockN{3\textsuperscript{rd} Mi Lu}
		\IEEEauthorblockA{\textit{Electrical and Computer Engineering} \\
			\textit{Texas A\&M University}\\
			College Station, TX \\
			mlu@ece.tamu.edu}
	}
	
	\maketitle
	
	\begin{abstract}
		Dark Channel Prior (DCP) is a widely recognized traditional dehazing algorithm. However, it may fail in bright region and the brightness of the restored image is darker than hazy image. In this paper, we propose an effective method to optimize DCP. We build a multiple linear regression haze-removal model based on DCP atmospheric scattering model and train this model with RESIDE dataset, which aims to reduce the unexpected errors caused by the rough estimations of transmission map $t(x)$ and atmospheric light $A$. The RESIDE dataset provides enough synthetic hazy images and their corresponding groundtruth images to train and test. We compare the performances of different dehazing algorithms in terms of two important full-reference metrics, the peak-signal-to-noise ratio (PSNR) as well as the structural similarity index measure (SSIM). The experiment results show that our model gets highest SSIM value and its PSNR value is also higher than most of state-of-the-art dehazing algorithms. Our results also overcome the weakness of DCP on real-world hazy images.
	\end{abstract}
	
	
	\section{Introduction}
	Images captured in outdoor scenes are usually degraded by haze, fog and smoke. Suffering from poor visibility, reduced contrasts, fainted surfaces and color shift, hazy images will miss many details. As a result, most outdoor vision applications such as video-surveillance systems, traffic monitoring systems and object detection systems fail to work normally in some cities often covered with haze. Thus, haze removal is highly desired in computational photography and computer vision applications. However, the existence of haze adds complicated nonlinear and data-dependent noise to the images, making the haze removal a highly challenging image restoration and enhancement problem. Many attempts have been made to recover the haze-free images from hazy images \cite{b15} \cite{b16} \cite{b17} \cite{b18}. Among traditional dehazing algorithms, the dark channel prior (DCP) \cite{b1} is widely recognized. 
	
	DCP is a regular pattern found in general haze-free images. In most non-sky patches of the haze-free image, at least one color has some pixels whose intensity are very low and close to zero. The dehazing algorithm based on DCP is simple and effective. That's why it becomes popular as soon as it is proposed. However, the traditional dark channel prior has a rough estimation on the transmission $t(x)$ and the Atmospheric Light \textbf{A}. This weakness is very obvious when the scene object is inherently similar to the air light over a large local region and no shadow is cast on it, the brightness of the restored image will be darker than real haze-free image. And when there is a bright region (sky etc.), the strong sunlight will lead to color distortion and shift in restored images. Many researchers have tried various methods to make up this weak point, which will be discussed in Section III. In this paper, we propose an improved multiple linear regression haze-removal model to optimize the accuracy of estimation on the transmission t(x) and the Atmospheric Light \textbf{A}. And the haze-removal model is trained with the training set of REalistic Single Image DEhazing (RESIDE) dataset \cite{b3}. We also evaluate our algorithm based on the test set of RESIDE. Experiment results show that our improved model achieves the highest SSIM value campared with all the state-of-the-art algorithms. Finally, PSNR of our method is promoted by 5.3 compared to DCP rule and SSIM is promoted by 0.23. We also compared the recovered images from our algorithm with the recovered images from DCP, the improvement of our algorithm is obvious and significant.
	
	In Section II, we will discuss over the principle and limitation of DCP. In Section IV, we will introduce our method to improve DCP in detail. The experiment results will be shown in Section V. 
	
	\section{Background} \label{Background}
	In computer vision, the classical model to describe the generation of a hazy image is the atmospheric scattering model:
	\begin{equation} \label{eq:1}
	\textbf{\textit{I}}(x) = \textbf{\textit{J}}(x)\textit{t}(x) + \textbf{\textit{A}}(1 - \textit{t}(x))
	\end{equation}
	where $I(x)$ is the observed intensity, $J(x)$ is the scene radiance, $A$ is the atmospheric light, and $t(x)$ is the medium transmission matrix. The first term $J(x)t(x)$ on the right-hand side is called direct attenuation, and the second term $A(1-t(x))$ is called airlight.
	When the atmosphere is homogeneous, the transmission matrix $t(x)$ can be defined as:
	\begin{equation} \label{eq:2}
	\textit{t}(x) = \textit{e}^{-\beta d(x)}
	\end{equation}
	where $\beta$ is the scattering coefficient of the atmosphere, and $d(x)$ is the scene depth, which indicates the distance between the object and the camera.
	
	Given the atmospheric scattering model (\ref{eq:1}), most state-of-the-art single image dehazing algorithms estimate the transmission matrix $t(x)$ and the global atmospheric light $A$. Then they recover the clean images $J(x)$ via computing the transformation of (\ref{eq:1}):
	\begin{equation} \label{eq:3}
	\textbf{\textit{J}}(x) = \frac{1}{\textit{t}(x)}\textbf{\textit{I}}(x) - \textbf{\textit{A}}\frac{1}{\textit{t}(x)} + \textbf{\textit{A}}
	\end{equation}

	\section{Overview of existing algorithms}
	\subsection{Dehazing algorithm based on Dark Channel Prior}
	To formally describe the observation in DCP, the dark channel of an image $J$ is defined as:
	\begin{equation} \label{eq:4}
	\textbf{\textit{J}}^{dark}(x) = \min_{y\in \Omega (x)}(\min_{c} \textbf{\textit{J}}^{c}(y)) \approx 0 
	\end{equation}
	
	Then by minimizing both sides of equation (\ref{eq:1}), and putting (\ref{eq:2}) into (\ref{eq:1}), we can eliminate the multiplicative term and estimate the transmission $\widetilde{t}$ by:
	\begin{equation} \label{eq:5}
	\widetilde{t} = 1 - \min_{y\in \Omega (x)}(\min_{c} \frac{\textbf{\textit{J}}^{c}(y)}{\textbf{\textit{A}}^{c}})	
	\end{equation}
	
	To keep the vision perceptual well and looks natural, He et al\cite{b1} have introduced a constant parameter $w$ $(0 <  w \leq 1)$ to keep a very small amount of haze for the distant objects, then the transmission becomes:
	\begin{equation} \label{eq:6}
	\widetilde{t} = 1 - w\min_{y\in \Omega (x)}(\min_{c} \frac{\textbf{\textit{I}}^{c}(y)}{\textbf{\textit{A}}^{c}})	(4)
	\end{equation}
	
	When estimating   $A$, DCP also takes the sunlight into consideration based on Tan's work \cite{b9}. He at el\cite{b1} have adopted the dark channel to detect the most haze-opaque region and improve the atmospheric light estimation. They first pick the top 0.1 percent brightest pixels in the dark channel. These pixels are usually for the most haze-opaque region. Among these pixels, the pixels with the highest intensity in the input image $I$ are selected as the atmospheric light.
	
	\subsection{Limitations of DCP and improved algorithms}
	However, DCP has many limitations when estimating $t(x)$ and $A$, which make DCP fail in some specific cases that are very common in real world. When the scene object is inherently similar to the air light (e.g., snowy ground or a white wall) over a large local region and no shadow is cast on it, it will underestimate the transmission of these objects and overestimate the haze layer. So the brightness of the restored image will be darker than the real haze-free image. 
	
	Transmission map $t(x)$ estimation is one of the most important parts in state-of-the art dehazing algorithms. Due to the rough estimation of $t(x)$ in DCP, some algorithms are proposed to optimize the estimation of $t(x)$. He et al\cite{b1} have fixed $w$ to 0.95 for all results in their paper, which intends to remove most haze and leave a small amount of haze for human to perceive depth. However, Chen et al\cite{b4} have pointed out that it sometimes causes haze removed too much, image with only a little haze often leads to an unreal feeling of vision. Thus, they proposed an dehazing parameter adaptive method to change $w$ corresponding to haze distributions. The dehazing parameters are estimated locally instead of globally. The original DCP uses soft matting to refine the estimation of transmission. Later they propose guide filter \cite{b6} to optimize the rough transmission $t(x)$, which turns out both more effective and more efficient than soft matting. And the traditional dark channel prior also has not fully exploited its power due to improper assumptions or operations, which causes unwanted artifacts. Then Zhu at el\cite{b5} have introduced a novel method for estimating transmission $t(x)$ by energy minimization to solve this problem. And several attempts have been made to improve the restoration of sky region \cite{b10} \cite{b11} \cite{b12}.
	
	As for the limitations of estimation of atmospheric light $A$, He et al\cite{b1} consider the sunlight and add it to the scene radiance of each color channel:
	
	\begin{equation} \label{eq:7}
	\textbf{\textit{J}}(x) = \textbf{\textit{R}}(x)(\textbf{\textit{S}} + \textbf{\textit{A}})
	\end{equation}
	where $R \leq 1$ is the reflectance of the scene points. Then We put (\ref{eq:7}) to (\ref{eq:1}):
	\begin{equation} \label{eq:8}
	\textbf{\textit{J}}(x) = \textbf{\textit{R}}(x)\textbf{\textit{S}}t(x) + \textbf{\textit{R}}(x)\textbf{\textit{A}}t(x) + (1 - t(x))\textbf{\textit{A}}
	\end{equation}
	
\begin{figure}
		\subfigure[Hazy image]
		{
			\includegraphics[width=1.5in]{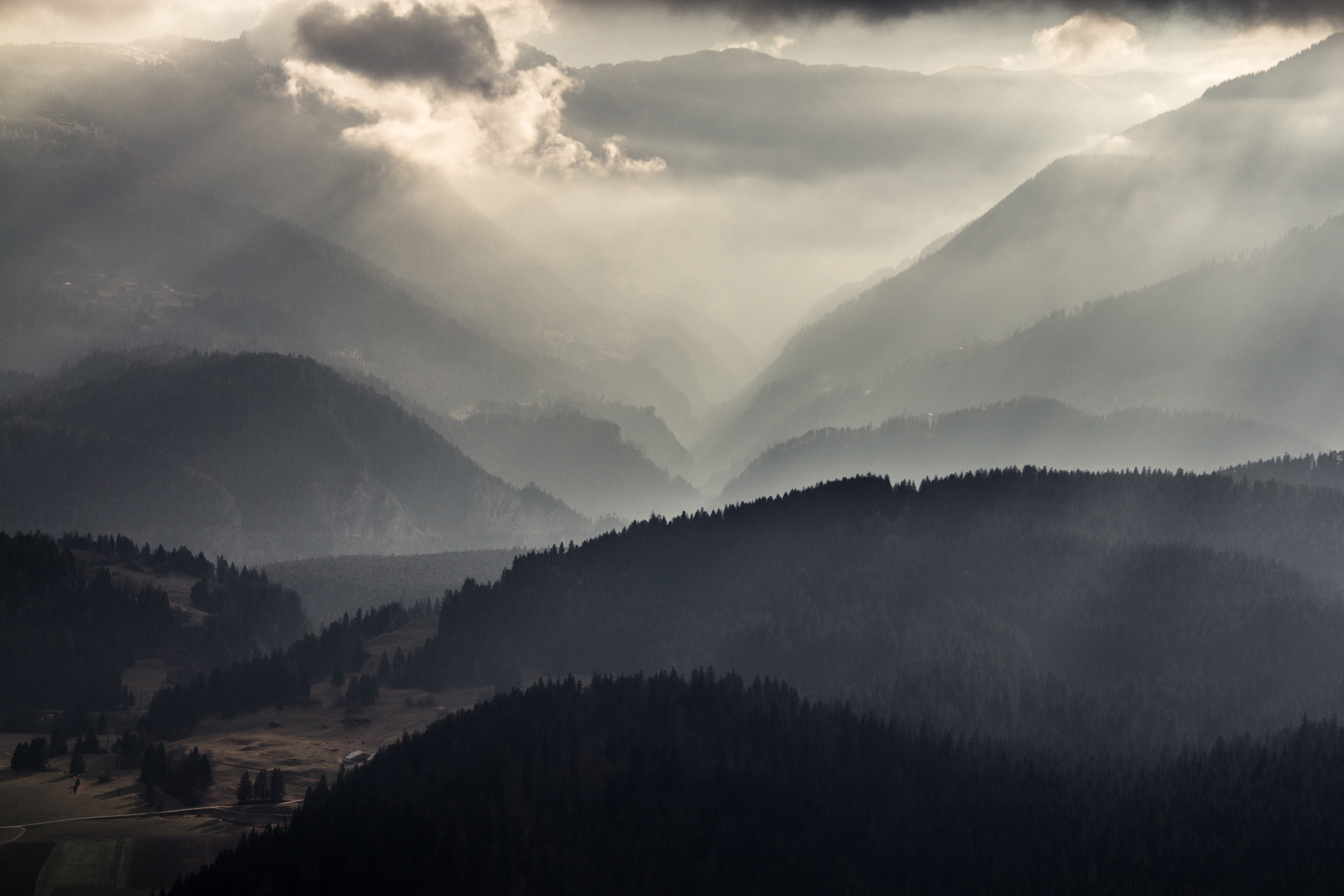}
			\label{b_8}
		}
		\subfigure[Recovered image via DCP]
		{
			\includegraphics[width=1.5in]{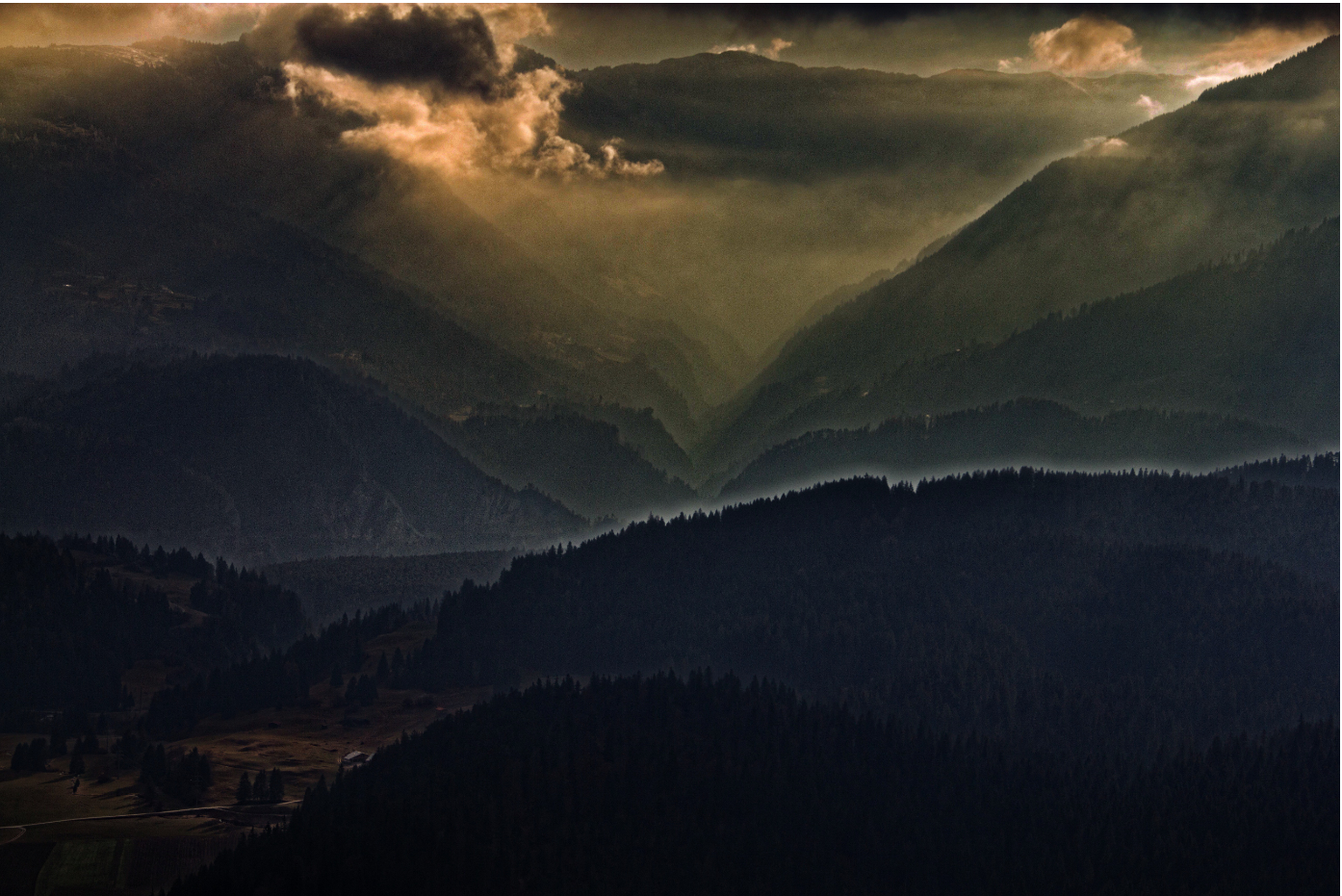}
			\label{b_9}
		}
		\caption{The weakness of original DCP}
		\label{fig:The color distortion of DCP in sky region}
	\end{figure}

	Although taking sunlight into account, He at el \cite{b1} still think the influence of sunlight is weak due to small $t(x)$. However, DCP restricts the transmission map $t(x)$ by a lower bound $t_{0}$ in case the recovered scene radiance $J(x)$ is prone to noise:
	\begin{equation} \label{eq:9}
	\widetilde{t} = \max(t(x), t_{0})
	\end{equation}
	where $t_{0} = 0.1$. Since the lower bound of $t(x)$ is 0.1, the influence of sunlight can be significant in the sky region with strong sunlight. We compare the hazy image with the recovered image from DCP in Figure \ref{fig:The color distortion of DCP in sky region}, and the color distortion in the sky region can be obviously observed.
	
	\subsection{Overview of CNN methods}
	With the development of Convolutional Neural Network (CNN) in computer vision problems \cite{b19} \cite{b20} \cite{b21} \cite{b22}, some deep learning methods have been applied to solve haze-removal problem. A multi-scale deep neural network is introduced in \cite{b14}. It generates coarse-scale net to predict a holistic transmission map $t(x)$ and refine results locally with a fine-scale net. Cai et al\cite{b13} have proposed a trainable end-to-end model called DehazeNet to estimate meiudm transmission. DehazeNet takes hazy iamges as imputs and estimates transmission map $t(x)$. It also proposes a novel nonlinear activation function to improve the quality of recovered haze-free image. Based on the RESIDE dataset, in \cite{b2}, an image dehazing convolutional neural network called All-in-One Dehazing Network (AOD-Net) is built. AOD-Net directly generates the clean image through a light-weight CNN. And as an end-to-end model, it's easy to be embedded into other deep models, to improve high-level tasks on hazy images. 
	
	One severe problem of state-of-the-art dehazing algorithms is about the way they compare the performances of different algorithms. Many researchers just list the original images and their recovered images together with the haze-removed images of other algorithms. They expect the readers to rank the performances of these images through observation. This is not a convincing way to prove that the new algorithm performs better than previous algorithms, especially when the difference is not very obvious. Wang in \cite{b3} has created a new large-scale benchmark consisting of both synthetic and real-world hazy images, called RESIDE. The dataset provides a rich variety of criteria for dehazing algorithm evaluation, ranging from full-reference metrics, to no-reference metrics, to subjective evaluation and the novel task-driven evaluation, which means the performances of different dehazing algorithms can be obviously compared via a variety of criteria. Wang in \cite{b3} has also tested other dehazing algorithms including state-of-the-art methods and deep learning methods on the test set of RESIDE, which makes it possible to compare all dehazing algorithms directly via the comparison of criteria. However, the performances based on the criteria can not reflect the limitations of dehazing algorithms. From the criteria, we are not able to judge if DCP fails in the sky region. In Section V, we will compare the performances both from recovered images and two important criteria SSIM and PSNR. The PSNR value corresponds to the image quality. A higher PSNR value provides a higher image quality. And the SSIM is used to measure the similarity between two images \cite{b7}. It was developed by Wang et al\cite{b8} and it is considered to be correlated with the quality perception of the human visual system (HVS). 
	
	\section{Multiple linear regression haze-removal model}
	After the transmission map $t(x)$ and the Atmospheric Light $A$ are estimated, hazy image can be recovered by:
	
	\begin{equation} \label{eq:10}
	\textbf{\textit{J}}(x) = \frac{\textbf{\textit{I}}(x)}{\textit{t}(x)} - \frac{\textbf{\textit{A}}}{\textit{t}(x)} + \textit{A}
	\end{equation}
	
	In Section II, we already introduced the estimation of transmission map $t(x)$ and the Atmospheric Light $A$ are both based on the hazy images. Although many algorithms attempt to refine the estimations on transmission map $t(x)$, and the Atmospheric Light $A$, estimations can still generate some unexpected deviations. And this kind of errors are normally impossible to eliminate. Now we introduce the multiple linear regression model to optimize the atmospheric scattering model (\ref{eq:10}). Multiple linear regression is the most common form of linear regression analysis. As a predictive analysis, multiple linear regression is used to explain the relationship between one continuous dependent variable and two or more independent variables. When we train with thousands of synthetic images and haze-free images, the scene radiance $J(x)$, which is also the RGB pixel of haze-free image, can be regarded as the the continuous dependent variable.
	For each image, after we estimate $t(x)$ and $A$ by original DCP,  $\frac{\textbf{\textit{I}}(x)}{\textit{t}(x)}$, $\frac{\textbf{\textit{A}}}{\textit{t}(x)}$ and $A$ in (\ref{eq:7}) can be regarded as three independent variables, where $I(x)$ is the pixel of hazy image. Then both two parameters $t(x)$ and $A$ together with the pixels of input images and target images can be simplified to a prediction problem with multiple linear regression model, which describes how the mean response $J(x)$ changes with the three explanatory variables:
	
	\begin{equation} \label{eq:11}
	\textbf{\textit{J}}(x) = w_{0} \frac{\textbf{\textit{I}}(x)}{\textit{t}(x)} + w_{1} \frac{\textbf{\textit{A}}}{\textit{t}(x)} + w_{2} \textbf{\textit{A}} + b
	\end{equation}
	
	Then we implement Stochastic Gradient Descent (SGD) to our multiple linear regression model, which is one of the most widely used algorithms to solve optimization problems. The Outdoor Training Set (OTS) of RESIDE dataset \cite{b3} provides 8970 outdoor haze-free images. For each haze-free image, OTS also provides about 30 synthesis hazy images with haze intensity from low to high. Different haze intensity means a lot to our multiple linear regression haze-removal model, because we don't expect that our algorithm can only perform good on hazy images with fixed haze intensity. When we train our model on OTS, we refer haze-free images as target images $J$, refer synthesis images as input images $I$, take the images recovered from our model as output images $J_{\omega}$. Then (\ref{eq:11}) can be re-formulated as:
	\begin{equation} \label{eq:12}
	\textbf{\textit{J}}_{\omega}(x) = w_{0} \textbf{\textit{x}}_{0} + w_{1} \textbf{\textit{x}}_{1} + w_{2} \textbf{\textit{x}}_{2} + b
	\end{equation}
	For that $I(x)$ and $A$ are both defined based on RGB color channels, the dimension of three weights and bias is (3,1), aiming to refine the parameters from all three color channels. 
	And the deviations of the output images $J_{\omega}$ from the target images $J$ are estimated by mean-squared error (MSE):
	\begin{equation} \label{eq:13}
	\textbf{\textit{MSE}} = (\textbf{\textit{J}} - \textbf{\textit{J}}_{\omega})^{2}
	\end{equation}
	
	By training our model, we try to find the optimal weights and bias that minimize the mean-squared error.
	\begin{itemize}
		\item Define the cost function of our model based on (\ref{eq:13}):
		\begin{equation} \label{eq:14}
		\textbf{\textit{L}}(\omega) = \frac{1}{2n} \sum_{i=1}^{n} (\textbf{\textit{J}}^{(i)} - \textbf{\textit{J}}_{\omega}^{(i)})^{2}
		\end{equation}
		\item For each image in training set, we repeatedly update three weights via:
		\begin{equation} \label{eq:15}
		\omega_{k} = \omega_{k} - \alpha \frac{\partial }{\partial \omega} \textbf{\textit{L}}(\omega)
		\end{equation}
		where $\alpha$ is the learning rate, and $k \in {0,1,2}$. 
		\item Since the derivative of cost function can be computed via (\ref{eq:14}), Equation (\ref{eq:15}) can be simplified as:
		\begin{equation} \label{eq:16}
		\omega_{k} = \omega_{k} - \alpha \frac{1}{n} \sum_{i=1}^{n} (\textbf{\textit{J}}^{(i)} - \textbf{\textit{J}}_{\omega}^{(i)})x_{k}
		\end{equation}
		\item For each image in training set, we repeatedly update the bias via:
		\begin{equation} \label{eq:17}
		b = b - \alpha \frac{1}{n} \sum_{i=1}^{n} (\textbf{\textit{J}}^{(i)} - \textbf{\textit{J}}_{\omega}^{(i)})
		\end{equation}
	\end{itemize}
	
	And after we get the optimal weights and bias, they can be simply added to traditional DCP. What's more, the multiple linear regression model will not modify the inner theory of any state-of-the-art dehazing algorithm. Our model can be applied to further improve the performance of some state-of-the-art dehazing algorithms that optimize the estimations of $t(x)$ and $A$. 
	
	Actually, the performance of our model can be further improved by providing higher quality synthetic hazy images. RESIDE dataset is good enough to overcome the weakness of traditional DCP. However, the synthesis images can only generate average haze, which are very different from real-world hazy images. So our model can't fully remove haze when the haze intensity is high in part of the image. And RESIDE dataset does not include the images of night scenes. Without training on night scenes images, our algorithm fails to recover the hazy images with dark light.

	\section{Results} 
	In this section, by demonstrating our dehazing results on several groups of hazy images as well as comparing the PSNR and SSIM value, we show the better performance of our model over DCP and other dehazing algorithms.
	
	\subsection{Experiment setup}
	Most recently, a benchmark dataset of both synthetic and real-world hazy images provided in \cite{b3} for dehazing problems are introduced to the community. In our experiment, the Synthesis Object Testing Set (SOTS) in RESIDE dataset is used to test the dehazing performance of our method. Our algorithm focuses on recovering outdoor hazy images, therefore, we won't evaluate the performance on recovering the 500 synthetic indoor images in SOTS.
	
	In the training phase, 8000 haze-free images and corresponding synthetic hazy images in Outdoor Training Set (OTS)\cite{b3} are used to train our multiple linear regression haze-removal model. In the testing phase, 500 outdoor haze-free images and 500 outdoor synthetic hazy images in Synthetic Objective Testing Set(SOTS) are tested using the well trained model. We compare the performance with original DCP from both the average SSIM and PSNR, and we also list the comparison of recovered images between our algorithm and DCP to prove that the multiple linear regression haze-removal model can overcome the weakness of DCP.

	\subsection{Experiment results on SOTS}
\begin{figure}
		\subfigure[Synthetic hazy image]
		{
			\includegraphics[width=1.6in]{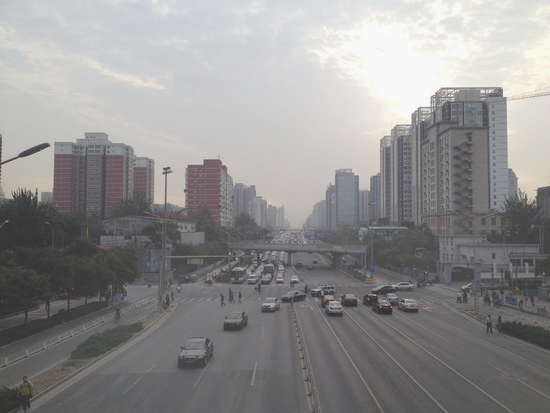}
			\label{b_11}
		}
		\subfigure[Haze-free image]
		{
			\includegraphics[width=1.6in]{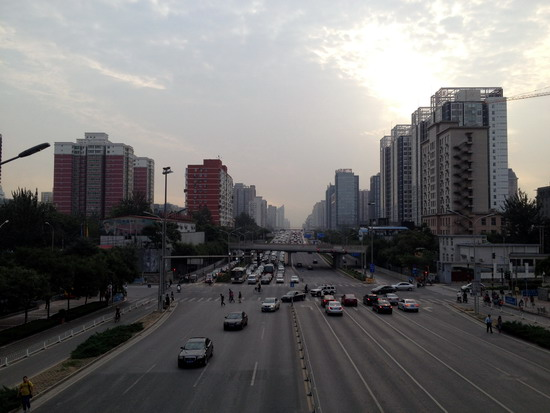}
			\label{b_12}
		}
		\\
		\subfigure[Recovered image by DCP]
		{
			\includegraphics[width=1.6in]{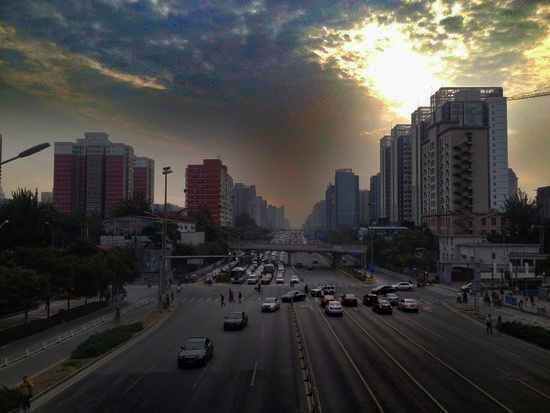}
			\label{c_13}
		}
		\subfigure[Recovered image by our model]
		{
			\includegraphics[width=1.6in]{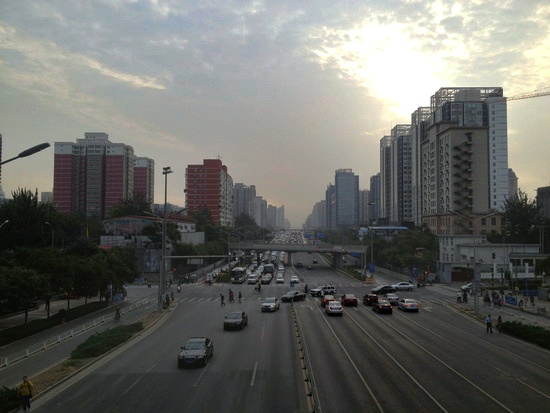}
			\label{d_13}
		}
		\caption{Comparison on synthetic hazy image}
		\label{fig:Comparison on synthetic hazy image}
	\end{figure}

	In Fig. \ref{fig:Comparison on synthetic hazy image}, We demonstrate the performance of multiple linear regression haze-removal model from the difference observed through recovered images via original DCP and our improved model. Fig. \ref{fig:Comparison on synthetic hazy image} (a) shows a synthetic hazy image selected from SOTS, and Fig. \ref{fig:Comparison on synthetic hazy image} (b) shows its haze-free image. Fig. \ref{fig:Comparison on synthetic hazy image} (c) shows the recovered image via original DCP. We can easily observe that the color distortion in sky region and non-sky region performs darker than its haze-free image. Fig. \ref{fig:Comparison on synthetic hazy image} (d) shows the recovered image via our model. Compared with the haze-free image, the haze is almost completely removed and the sky region seems more natural. The high similarity is obvious through observation.
	
	\begin{table}[htbp]
		\begin{center}
			\resizebox{1\columnwidth}{!}{
				\begin{tabular}{|c|c|c|}
					\hline
					\multicolumn{3}{|c|}{\textbf{500 outdoor image}}   \\ 
					\hline
					\textbf{Dehazing method name}  & \textbf{PSNR}    & \textbf{SSIM} \\
					\hline
					\textbf{Improved DCP model}  &   23.84  &  \textbf{0.9411} \\
					\hline
					\textbf{DCP}  &	 18.54	&	0.7100 \\
					\hline
					\textbf{FVR}  &   16.61  & 0.7236 \\
					\hline
					\textbf{BCCR}  &   17.71  &  0.7409 \\
					\hline
					\textbf{GRM}  &  20.77   &  0.7617 \\
					\hline
					\textbf{CAP}  &  23.95   &  0.8692 \\
					\hline
					\textbf{NLD}  &  19.52   &  0.7328 \\
					\hline
					\textbf{DehazeNet}  &   \textbf{26.84}  & 0.8264 \\
					\hline
					\textbf{MSCNN}  &   21.73  & 0.8313 \\
					\hline
					\textbf{AOD-Net}  &  24.08   & 0.8726 \\
					\hline
			\end{tabular}}
		\end{center}
		\caption{Average SSIM and PSNR comparison between different dehazing methods on 500 outdoor synthetic image in SOTS.}
		\label{tb:outdoor_results}
	\end{table}
	
	In Table \ref{tb:outdoor_results}, we compare our results with several state-of-the-art method with average SSIM and average PSNR. As we mentioned in Section III, PSNR and SSIM together can compare the image quality and similarity between recovered images and haze-free images. Since we want to apply our improved model on traffic dehazing problems, we only compare the results on 500 outdoor images in SOTS in this table. All the results of other algorithms come from \cite{b3}. As shown, our multiple linear regression haze-removal model has improved PSNR by $5.3$ and SSIM by $0.23$ compared to the original DCP. When compared with well-recognized state-of-the-art dehazing algorithms and deep learning methods, our proposed model achieves reasonably good PSNR value and obtains the highest SSIM value. The results show the effectiveness of our model on dehazing outdoor images.

	\subsection{Dehazing results of realistic hazy images}
\begin{figure}
		\subfigure[Hazy image (1)]
		{
			\includegraphics[width=1.6in]{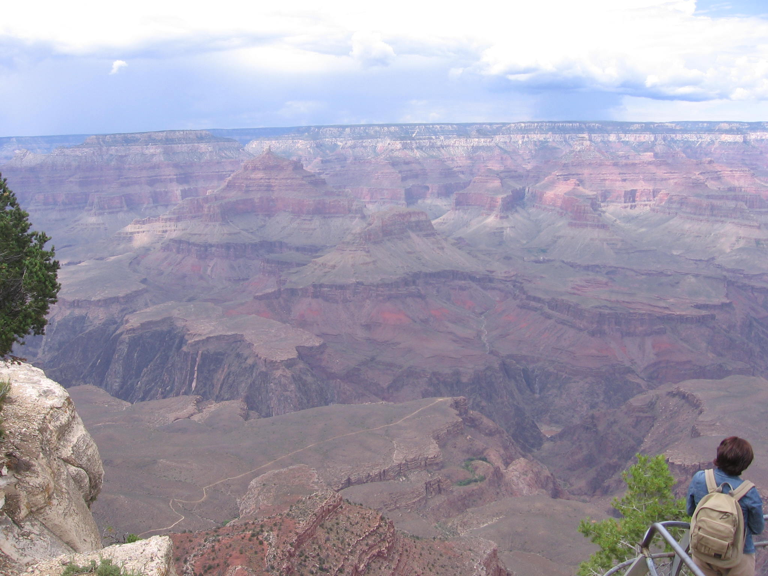}
			\label{b_1}
		}
		\subfigure[Hazy image (2)]
		{
			\includegraphics[width=1.6in]{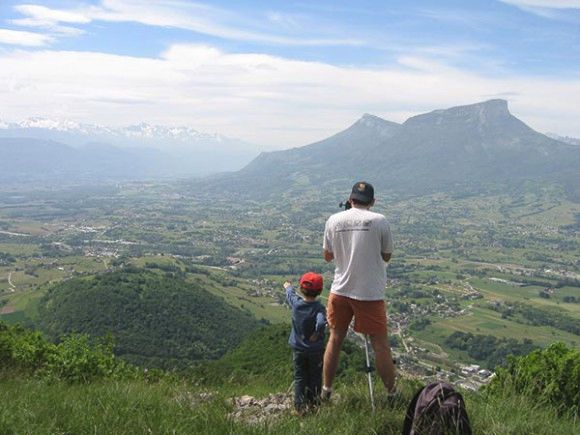}
			\label{b_2}
		}
		\\
		\subfigure[Recovered image (1) via DCP]
		{
			\includegraphics[width=1.6in]{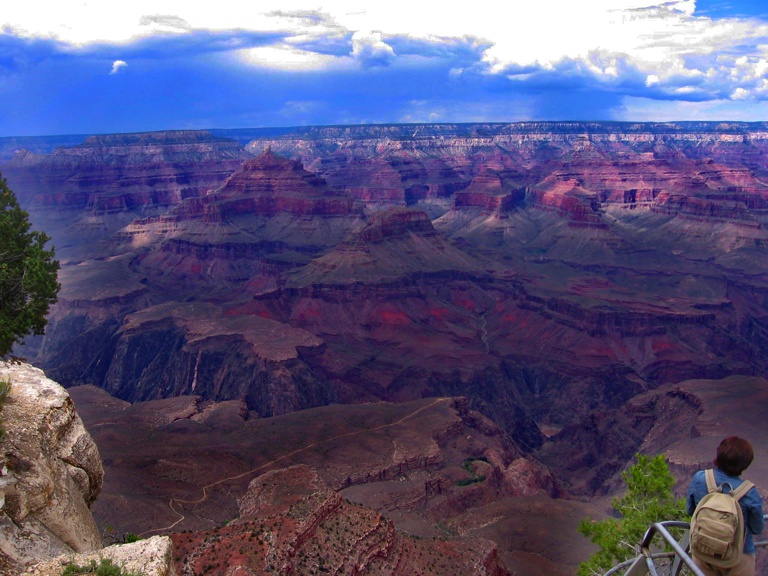}
			\label{c_1}
		}
		\subfigure[Recovered image (2) of via DCP]
		{
			\includegraphics[width=1.6in]{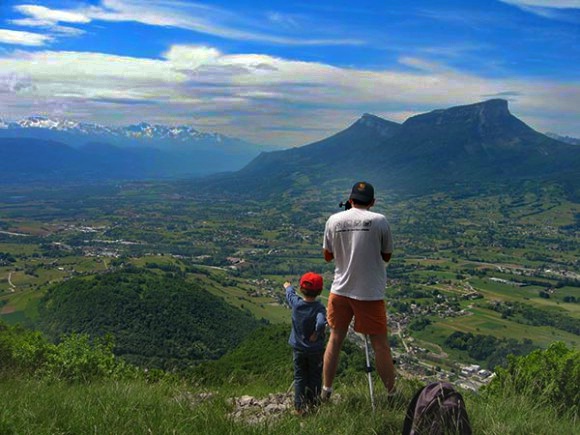}
			\label{d_1}
		}
		\\
		\subfigure[Recovered image (1) via our model]
		{
			\includegraphics[width=1.6in]{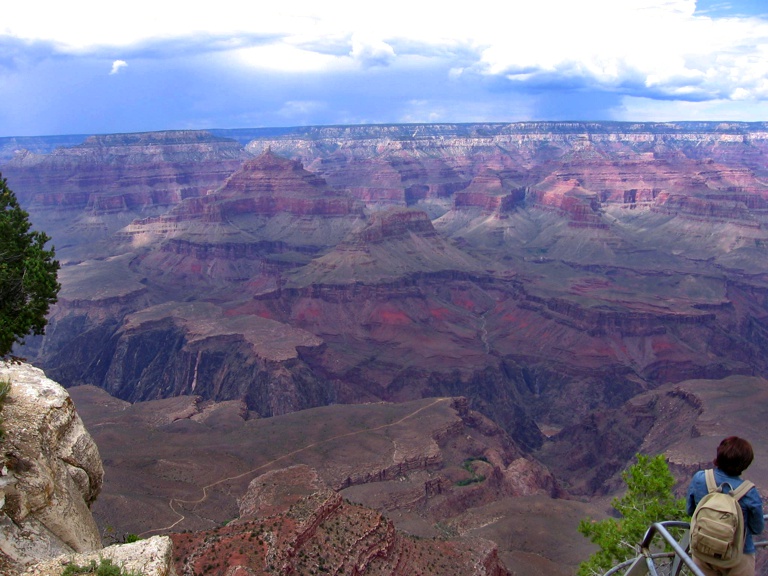}
			\label{c_2}
		}
		\subfigure[Recovered image (2) via our model]
		{
			\includegraphics[width=1.6in]{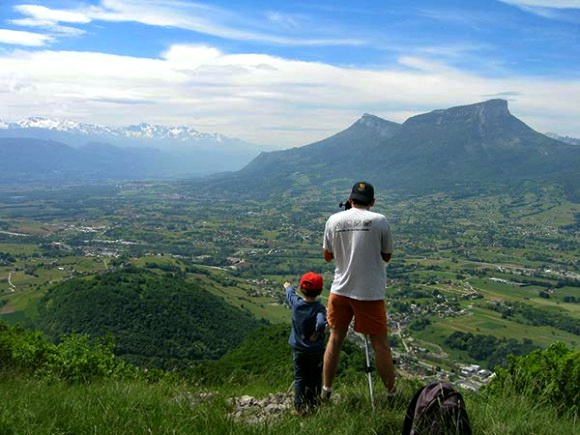}
			\label{d_2}
		}
		\caption{Comparison on real-world nature hazy images}
		\label{fig:real_image_nature}
	\end{figure}

\begin{figure}
		\subfigure[Hazy image (1)]
		{
			\includegraphics[width=1.6in]{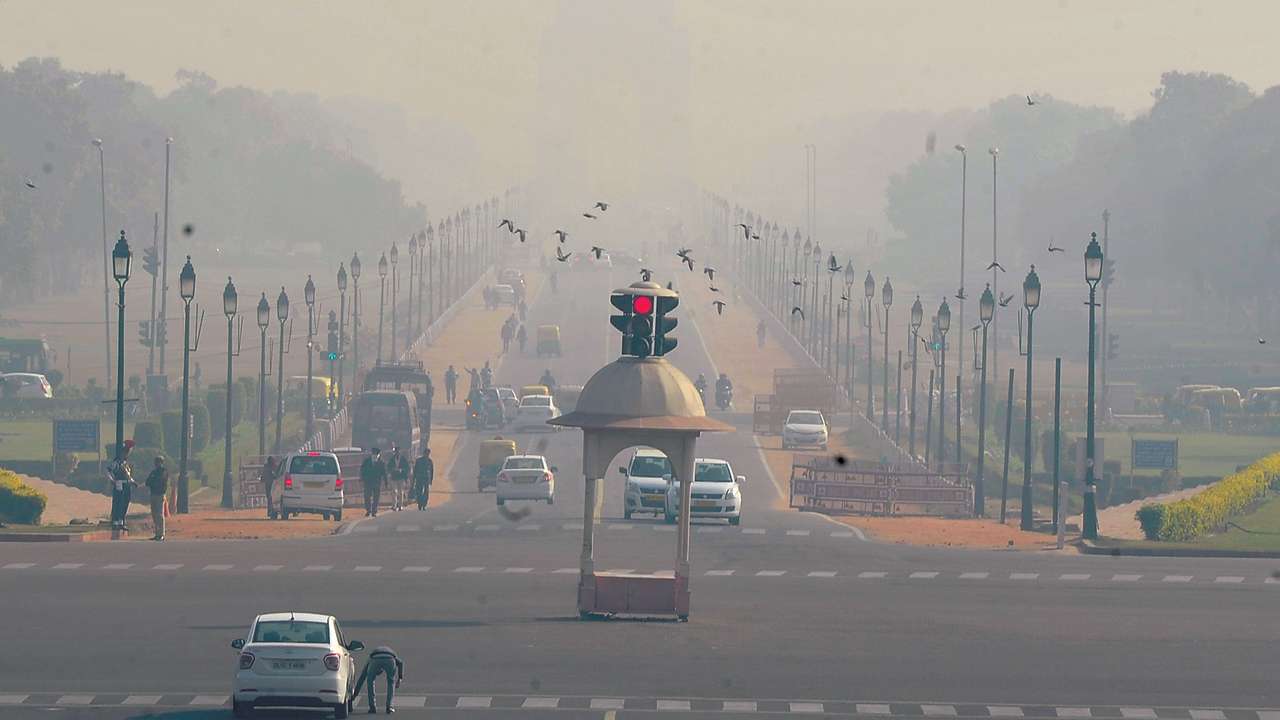}
			\label{b_4}
		}
		\subfigure[Hazy image (2)]
		{
			\includegraphics[width=1.6in]{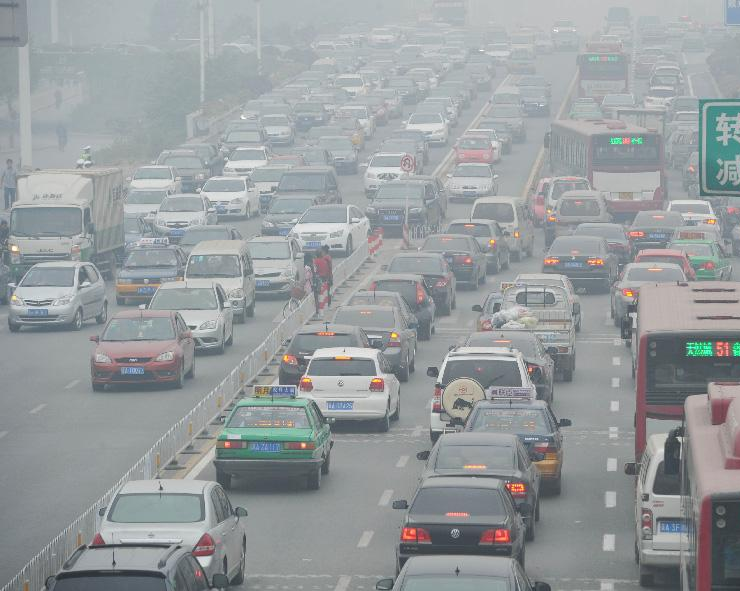}
			\label{b_5}
		}
		\\
		\subfigure[Recovered image (1) via DCP]
		{
			\includegraphics[width=1.6in]{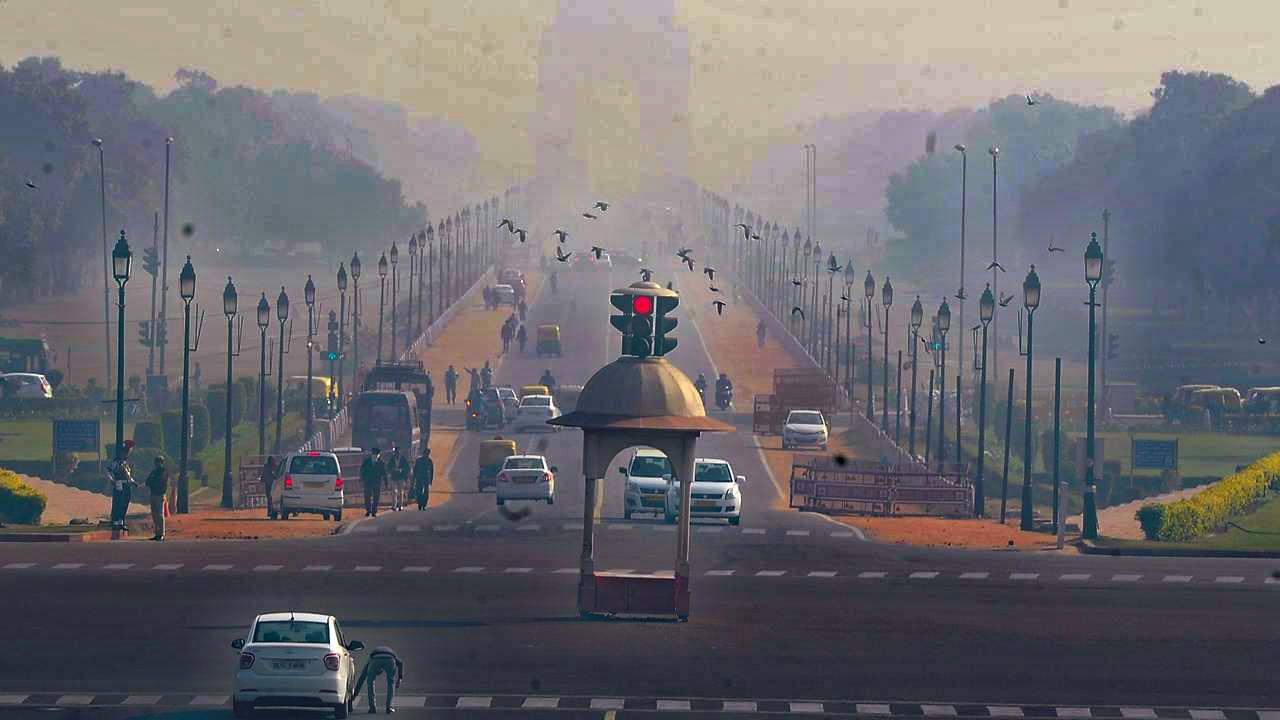}
			\label{c_4}
		}
		\subfigure[Recovered image (2) of via DCP]
		{
			\includegraphics[width=1.6in]{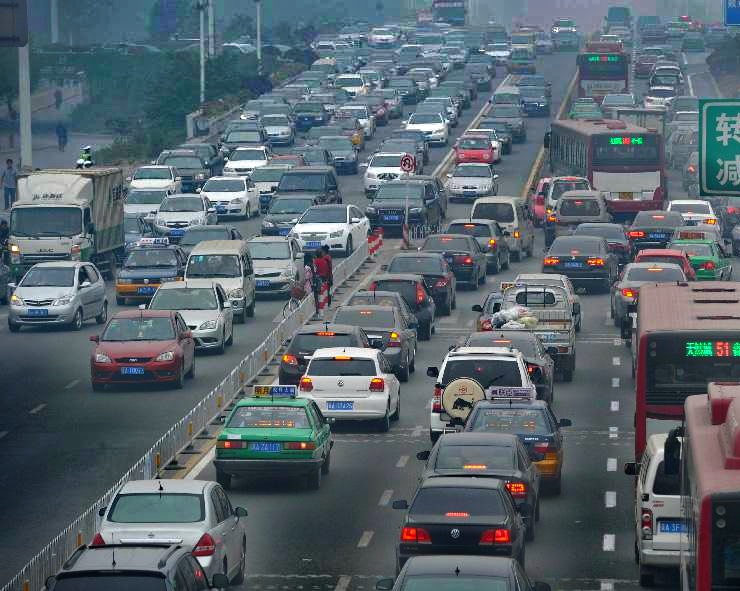}
			\label{d_4}
		}
		\\
		\subfigure[Recovered image (1) via our model]
		{
			\includegraphics[width=1.6in]{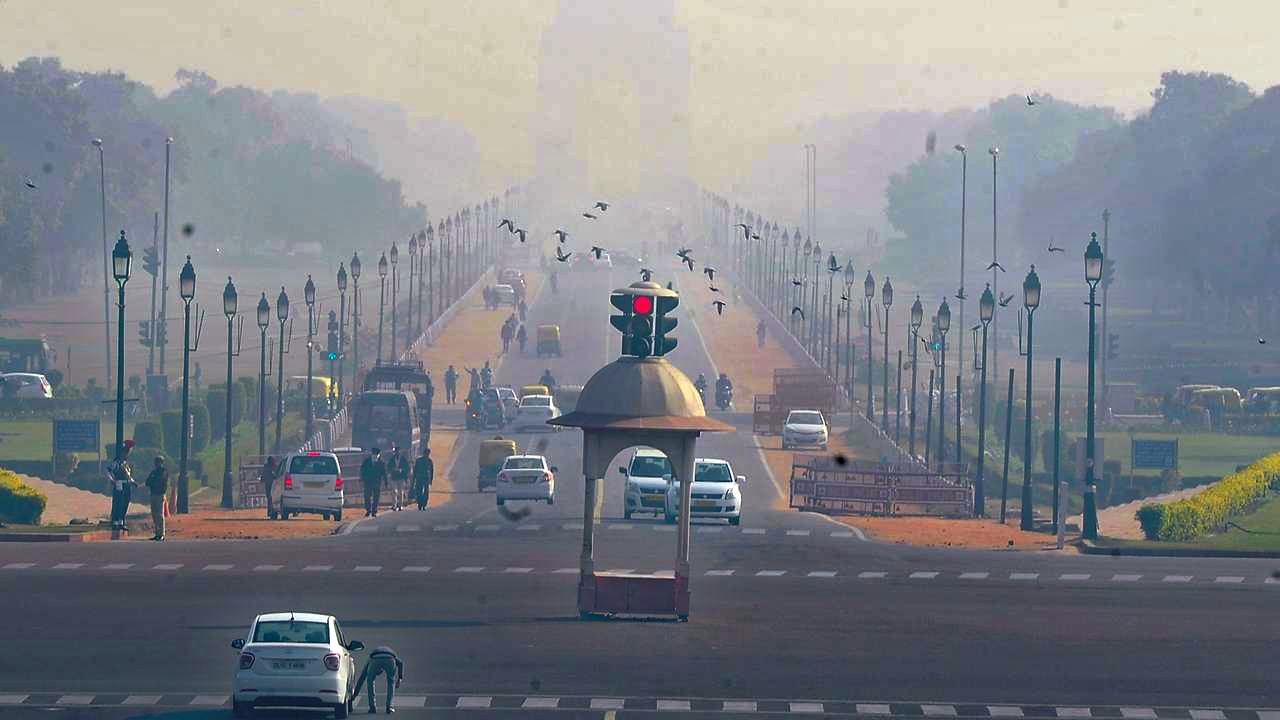}
			\label{c_5}
		}
		\subfigure[Recovered image (2) via our model]
		{
			\includegraphics[width=1.6in]{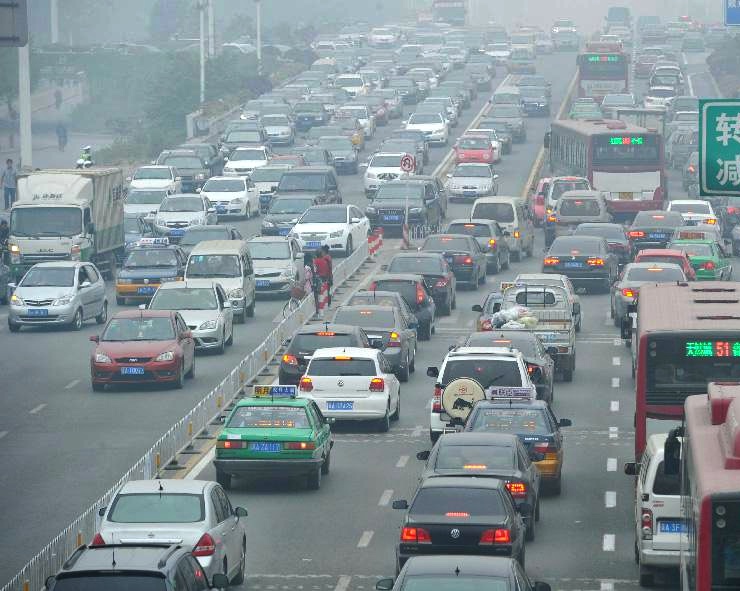}
			\label{d_5}
		}
		\caption{Comparison on real-world city hazy images}
		\label{fig:real_image}
	\end{figure}

	Fig. \ref{fig:real_image_nature} (a) and (b) present two commonly used hazy images and Fig. \ref{fig:real_image} (a) and (b) present two real-world hazy images in cities. Both Fig. \ref{fig:real_image_nature} and Fig. \ref{fig:real_image} show the comparison of recovered images via DCP and our model. Through observation, our model performs better on all these four images than original DCP. Our model prevents the color distortion in sky region and the brightness of whole image seems more natural than that of DCP. This advantage can help a lot when we apply dehazing algorithms to object detection in haze. Because the dark brightness of recovered images via DCP will lower the accuracy of detecting an object, such as car, in haze environment.
	
	\section{Conclusion}
	Because of the deficiencies of DCP we mentioned in Section II,  in this paper, we proposed a multiple linear regression haze-removal model to improve Dark Channel Prior. Two important parameters transmission map $t(x)$ and atmospheric light $A$ are estimated by original DCP. Weights and bias are added to the atmospheric scattering model to refine the estimation accuracy of $t(x)$ and $A$. As shown in Section V, our model not only achieved high values on PSNR and SSIM compared with several state-of-the-art algorithms, but also restores high quality dehazed images. And it can be easily combined with other models, making it possible to apply it on video dehazing and object detection in haze, which are both important in modern traffic.

\end{document}